\documentclass[10pt,twocolumn,letterpaper]{article}

\usepackage{cvpr}
\usepackage{times}
\usepackage{epsfig}
\usepackage{graphicx}
\usepackage{amsmath}
\usepackage{amssymb}
\usepackage{lipsum}

\usepackage{multirow}
\usepackage[table,xcdraw]{xcolor}

\usepackage[breaklinks=true,bookmarks=false]{hyperref}

\cvprfinalcopy 

\def\cvprPaperID{****} 

\begin{document}

\title{Context Tricks for Cheap Semantic Segmentation}

\author{Thanapong Intharah and Gabriel J. Brostow\\
University College London\\
Gower Street, London, United Kingdom\\
{\tt\small [t.intharah,g.brostow]@cs.ucl.ac.uk}
}

\maketitle

\begin{abstract}
   Accurate semantic labeling of image pixels is difficult because intra-class variability is often greater than inter-class variability. In turn, fast semantic segmentation is hard because accurate models are usually too complicated to also run quickly at test-time.
   Our experience with building and running semantic segmentation systems has also shown a reasonably obvious bottleneck on model complexity, imposed by small training datasets. We therefore propose two simple complementary strategies that leverage context to give better semantic segmentation, while scaling up or down to train on different-sized datasets.

As easy modifications for existing semantic segmentation algorithms, we introduce Decorrelated Semantic Texton Forests, and the Context Sensitive Image Level Prior. 
The proposed modifications are tested using a Semantic Texton Forest (STF) system, and the modifications are validated on two standard benchmark datasets, MSRC-21 and PascalVOC-2010. In Python based comparisons, our system is insignificantly slower than STF at test-time, yet produces superior semantic segmentations overall, with just push-button training.
\end{abstract}

\section{Introduction}
For many applications, such as navigation or robot-interaction, semantic segmentation of images needs to be both \emph{accurate} and \emph{fast} to be worthwhile. The environment can change more or less abruptly, but typically, many frames will have combinations of the same frequently co-occurring classes.  We leverage this persistence of context to improve pixel classification accuracy, given finite quantities of training data.

We build on the successful Semantic Texton Forest (STF)~\cite{shotton2008} approach, and enhance it through two main contributions. First, the Decorrelated Semantic Texton Forest (DSTF) is proposed as a variant to the STF that essentially preserves the original's efficiency. The DSTF  uses hierarchical clustering to decorrelate classes that have confusingly similar appearance for an STF. 
We further improve accuracy by introducing a Context Sensitive Image Level Prior (Context Sensitive ILP). Training this multi-label prior to account for the co-occurrence of classes proves to be very helpful and substantially better than the more typical multi-class training of ILPs. 

\section{Related Work}

Many semantic segmentation algorithms require carefully tuned models and/or fully connected Conditional Random Fields (CRFs) to produce accurate per-pixel labelings. Here, we review the most relevant such methods, as well as algorithms similar to ours that are close to state-of-the-art, but sacrifice some accuracy for improved test-time efficiency.\\

\noindent\textbf{General Algorithms}

We base our approach on the basic STF because we wish to leverage its low computational complexity, allowing for very fast implementations if needed. Shotton~\etal introduced STF as a component of their Bag of Semantic Textons (BoST)~\cite{shotton2008} model. BoST is one of the earliest algorithms to still be competitive in semantic segmentation challenges, appearing on the leaderboards of many semantic segmentation papers \cite{Alvarez2014,Gonfaus2010,Fidler2012}. BoST still outperforms other state of the art algorithms in some categories, as shown in Table~\ref{tab:allresult}, despite running in real-time. The two main components of the BoST model were i) use of the newly introduced STF, and ii) application of the Textonboost \cite{shotton2007} approach for encoding local context information, generating the BoST for each patch. We henceforth refer to the first component as STF, and to their combined approach as BoST. The STF is trained by growing extremely randomized trees with  raw pixel image patches as features. Leaf nodes store the class distributions of the image patches that reached that node. Although the whole STF process is considered very efficient at inference time, the STF by itself produces fairly low quality results, because raw pixel patches are often not expressive enough to be discriminative between classes. 
BoST improves the results dramatically, but with some computational overhead. 
Our proposed system modifies the STF by adding only a little overhead, but achieves significantly better performance compared with BoST~\cite{shotton2008}. 

One approach known for using an image level prior is \cite{Csurka2008}. 
Their overall system has a chain of stages: i) extracting patches and their low-level features, ii) constructing high-level features (Fisher vectors), iii) training (predicting, in test time) a class scoring unit using the high-level features, iv) assigning scores to a pixel and propagating scores to oversegmented regions, and finally v) integrating with their image level prior to refine their labels. 
Comparing to us, we skip (i) and (ii) which are bottle neck of the algorithm and use DSTF which very efficient because no feature extraction is required; their image level prior does not exploit co-occurrence statistics but model only the presence of each class individually.
By combining three simple components: local appearance scoring, context sensitive ILP, and location potential, we show that our method is more simple and performs better than \cite{Csurka2008} on the MSRC-21 dataset, 77\% to 65\% on average recall.

Another system that proposes a simple architecture is \cite{farabet2013} which devised a multi-scale Convolutional Neural Network (CNN) to extract features of a pixel for the scene labeling task. Their multi-scale CNN is designed to capture different levels of information, ranging from small region appearance, neighborhood context, and up to the global context of the image. The system remains simple by having only two components: pixel classification and simple post processing to smooth the classification result. 
However, the system required a specially designed model and careful parameter tuning at training time to get comparable result to the state of the art algorithms. It is therefore hard to re-build the training step and to test on different datasets. They also made a version for RGB-D data from indoor scenes~\cite{Couprie2013a}. In our approach, although we use CNN image level feature descriptors, we picked a general purpose feature generator \cite{Chatfield2014} that can be used out of the box without any parameter tuning. At training time, our system needs very little parameter tuning to achieve good results on different datasets.


\noindent\textbf{CRF based Algorithms}

CRFs are used in many semantic segmentation algorithms to regularize output labels. The Hierarchical Conditional Random Field (HCRF)~\cite{Ladick2009} uses different levels of quantization, from pixels to segments. They operate under the assumption that there is unlikely to be a single optimal quantization level that is suitable for all object categories. 

Beyond regularizing just neighboring pixels, several works model the relationships between all pixels. 
In the Dense CRF semantic segmentation of \cite{Krahenbuhl2011}, mean field approximation and Gaussian filtering is used to make inference in fully connected models practicable. Further, \cite{Alvarez2014} demonstrates that using a Dense CRF to infer all test images at once gives better results than inferring one image at a time. Our approach shows that a 4-connected neighbor CRF model can achieve results comparable to the fully connected model (both were tested on the MSRC-21 dataset).\\

\noindent\textbf{More Sophisticated Algorithms}

Co-occurrence statistics had been exploited in semantic segmentation systems to boost accuracy. The HCRF was improved further in~\cite{Ladicky2010} by incorporating a co-occurrence potential as per-image context information, into the CRF energy function. We propose a simpler system that also uses the co-occurrence statistics, but incorporates them in a different manner. Our system achieves comparable average recall scores to \cite{Ladicky2010}, without tuning our parameters per-dataset.

Gonfaus~\etal~\cite{Gonfaus2010} proposed another improvement to the HCRF, by adding a new consistency potential to the model, called the harmony potential. The harmony potential encodes all possible combinations of labels, allowing regions to have more than one class, which was a perceived limitation of HCRF models. Further, in \cite{Boix2011}, they introduced three more cues into the local unary potential to improve recall scores over their previous version. While certainly worthwhile, these algorithms, \eg~\cite{Boix2011,Gonfaus2010,Ladick2009,Ladicky2010}, achieve ever better results by adding complexity to their models. Our system maintains a very plain model, using a simple 4-connected CRF with potts pairwise potentials to encourage harmonization of the neighboring pixels. Our proposed simple system outperforms ~\cite{Gonfaus2010} on both average and global recall scores.

A sophisticated model was successfully demonstrated in \cite{Fidler2012}, where the problems of semantic segmentation, object detection, and scene classification were cast as one holistic CRF model. Their parameters are learned via a structured learning algorithm, and inference is accomplished by a convergent message-passing algorithm. The model exploits various cues, such as scene type, co-occurrence statistics, the shape and location of the object, and different quantization levels to boost the segmentation result. In contrast, our proposed system exploits some of these important cues as context, but integrates them together with a much simpler model, achieving accuracy that approaches that of the more sophisticated model. 

Most recently, CNNs have also been exploited in a more sophisticated semantic segmentation framework \cite{Hariharan2014}. They compute feature vectors for each proposed region using two CNN's, trained especially on bounding boxes and free-form versions of the region. Thereafter, the concatenated feature vectors are passed through a linear SVM classifier to get a per-class potential. The final class label for each region is assigned via non-maximum suppression. Although, the system performed very well on the PascalVOC 2012 dataset, it did so at the expense of algorithm complexity.

\section{Cheap Semantic Segmentation Model}
A good semantic segmentation algorithm should exploit different levels of information: local appearance, global appearance, the context of the scene, and location statistics of objects. We leverage this information with an emphasis on simplicity, so that both large and small training sets can be exploited, and for efficiency at test-time. 

\subsection{Local Appearance and a Classifier}
The cornerstone of visual understanding is having a class-covariant local appearance representation and a compatible classification model. 
Although superpixels and region-based methods encode neighborhood information, we opt to work on individual raw pixels, curtailing the need to select superpixel algorithms and parameters per dataset. 

The Shotton~\etal STF~\cite{shotton2008} is one of the simplest useful local appearance classifiers because it processes raw pixel values of a small image patch without constructing a separate feature descriptor. 
Shotton~\etal~\cite{shotton2008} also propose BoST, working in concert with an STF, as a significant contribution, because an STF has limited expressiveness and moderate classification accuracy in itself. The price of BoST's improved accuracy is its significant computational cost, so we proceed with just the STF model and representation.

\subsubsection{Learning from Confusion}
The first proposed contribution of this paper is to introduce the Decorrelated Semantic Texton Forest (DSTF), which is an improved version of the STF with only slight additional computational cost at test time. The DSTF emerged from our observation that very similar appearance patches can reach the same leaf node in an STF tree at training-time, even when they have different class labels. This problem occurrs in a significant minority of cases. Therefore, the DSTF is designed specifically to reduce the incidence of such high-entropy leaf-nodes. 

The DSTF assumes that such ``confused'' leaf nodes are populated with patches from \emph{distinct} scene types, or categories. 
To distinguish them, we add an upstream classifier to infer a scene category\footnote{The terms `scene category', `scene,' and `cluster' all refer to the same concept when we are explaining our algorithm.} for each input image. The inferred scene category dictates which single specialist STF should process that image.
We train a set of separate STF's, one for each scene category. 

\begin{figure}[htb]
\begin{center}
   \includegraphics[width=0.7\linewidth]{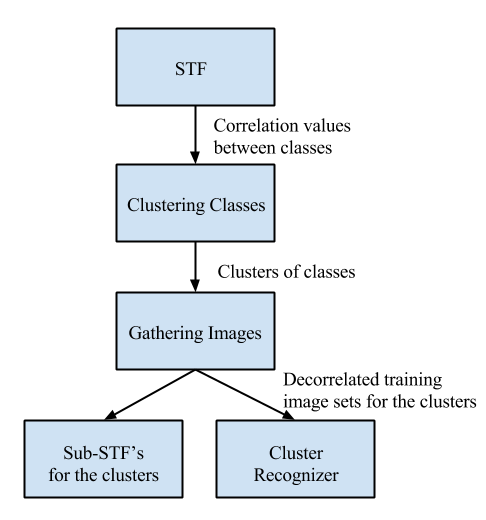}
\end{center}
\caption{Flow chart for training an upstream scene-category classifier (the Cluster Recognizer), and the downstream Decorrelated Semantic Texton Forest (DSTF) composed of multiple STF's, with one STF per scene category.}
\label{fig:dia}
\end{figure}

The scene categories are determined automatically, after growing a single temporary STF, depicted at the top of Figure~\ref{fig:dia}. The choice of categories we seek aims to group patches whose visual appearance does not confuse a single STF, and split visually similar patches from different classes that do confuse it. 




\noindent\textbf{Clustering Classes}
The temporary single STF is trained with all the images and classes. 
A label is associated with each pixel, which serves as the center of each small patch. In practice, we re-implement STFs of the original paper~\cite{shotton2007} and use the same parameter values, but without including their training invariance, because those parameters were not specified. 
Our early experiments showed that including some training invariance only improves the STF marginally.

Next, a class correlation matrix $\Omega$ is calculated by treating the class distributions at the leaf nodes of the trained STF as observations. Let $\mathbf{X}=\{X_1,...,X_T\}$ be the set of class distributions at the $T$ leaf nodes of the entire trained STF. ${X_i}=\{P(c_1|\mathbf{\mathcal{T}}_i),...,P(c_C|\mathbf{\mathcal{T}}_i)\}$ is a $C$-dimensional column vector of class probability at the leaf node $i$, conditioned on training examples $\mathbf{\mathcal{T}}_i$ that reached node $i$. The entries of class correlation matrix $\Omega$ are
\begin{equation}\label{omega}
\Omega(x,y)=\frac{\mathrm{Cov}(x,y)}{\sqrt{\mathrm{Cov}(x,x)*\mathrm{Cov}(y,y)}},
\end{equation}

\noindent where $\mathrm{Cov}(x,y)$ is the covariance between class $x$ and $y$ observed from the data $\mathbf{X}$ as row-slices across $\mathbf{X}$. We then cluster the semantic classes of the original problem by their correlation values, where the distance function is defined by 
\begin{equation}\label{dist}
\mathrm{Dist}(x,y)=\Omega(x,y)-\mathrm{min}(\Omega),
\end{equation}
and $\mathrm{Dist}(x,x)=0$. We subtract $\mathrm{min}(\Omega)$ from $\Omega(x,y)$ to make the smallest distance equal to zero. We group semantic classes by hierarchical clustering. 
To commit to hard cluster boundaries, we choose the minimum intra-cluster distance that forces every cluster to have at least three cluster members (classes), to prevent generating trivially small clusters.\\ 

\noindent\textbf{Gathering Images}
We use the class-clustering to gather the training images into new decorrelated (or less-correlated) training sets. 
We opt to gather images instead of patches to avoid overfitting. Gathering all images that contain class $c$ would risk piling almost all the training images into some ``specialist'' STF's, if a class is prevalent throughout, \eg sky. From experiments on our smallest dataset, MSRC-21's validation set, we found it already effective to use no more than the top-$7\%$ of all training images when training one of the cluster level STF's. 

The procedure to rank the images for each cluster follows. First, the class co-occurrence matrix $\Psi$ is computed from the ground truth images. Each element of $\Psi$ represents the probability of observing a class y given an image of class x, $\Psi(x,y)=P(y|I_x)$.  
Based on the matrix $\Psi$, we rank instances (images) for each class $c$ in the cluster separately, by assigning each image the score 
\begin{equation}\label{gather}
\mathrm{S}(c,G)=\sum_{i \in G}\Psi(c,\mathcal{L}(i)),
\end{equation}
where $i$ is a pixel in a ground truth image $G$, and $\mathcal{L}(\cdot)$ returns the ground truth label for the input pixel.

\noindent\textbf{Training Sub-STF's and the Cluster Recognizer}
From those decorrelated image training sets, we train separate standard STFs and the cluster recognizer. The cluster recognizer is a very fast linear SVM, trained on off-the-shelf CNN feature vectors~\cite{Chatfield2014}. At test time, the image is fed to the trained cluster categorizer that will redirect the image to an appropriate STF.

\begin{figure}[htb]
\begin{center}
   \includegraphics[width=1.0\linewidth]{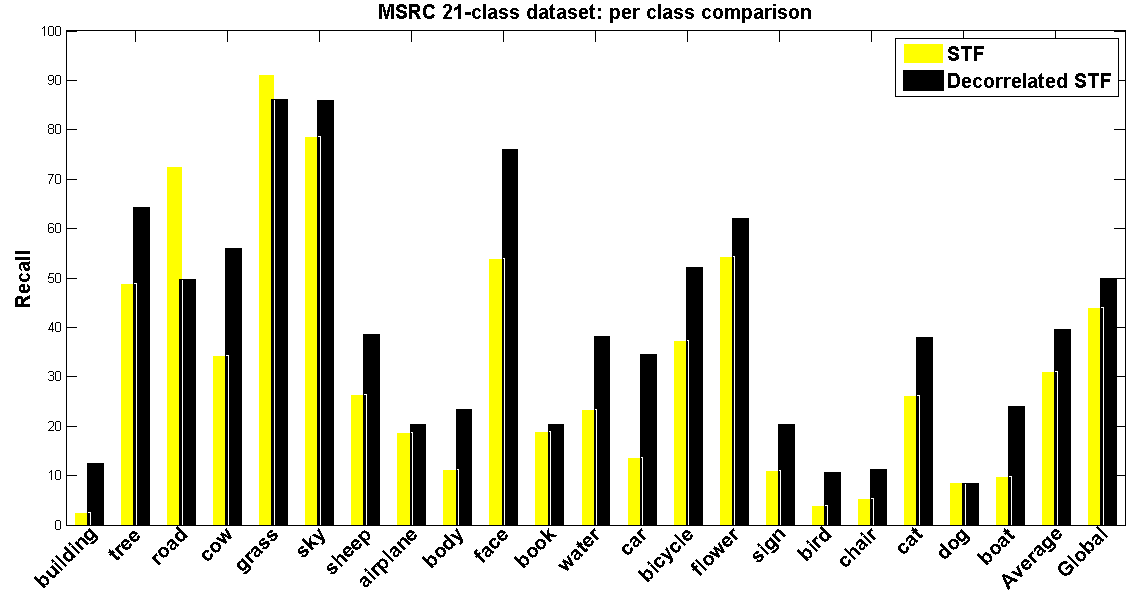}
\end{center}
   \caption{MSRC-21: Per-class comparisons of the DSTF results and STF results, measuring Recall rates.}
\label{fig:DSTF}
\end{figure}

A per-class comparison on the MSRC-21 dataset between normal STF and our DSTF is demonstrated in Figure~\ref{fig:DSTF}. From the figure, we can see that DSTF improves the segmentation accuracies for almost every class. Further analysis is deferred until Section \ref{ilp}.\\


\noindent\textbf{Efficiency Analysis}
We compare the efficiency of our proposed DSTF against BoST~\cite{shotton2008} because BoST is known to be a real-time semantic segmentation system that can be run on a standard 2008 PC. For simplicity, we count one inference computation of a decision tree or an SVM as one operation. Our goal is to compare the number of operations at test time. For \cite{shotton2008}, each pixel is routed through another randomized decision forest. At each node, the BoST for a related region of size $R$ pixels is constructed by inferring the STF for every pixel in the region. Hence, the number of operations required for predicting one pixel is $O(lR)$ where $l$ is the number of levels of the randomized decision forest. Whereas the DSTF requires inference by one scene-classifier, a linear SVM, to route the pixel to an appropriate STF, then inference by an STF for that pixel. Therefore, at test time, the DSTF spends merely two operations for one pixel prediction.

\subsection{Global Appearance and Objects Co-Occurrence}
Experments from previous work~\cite{Boix2011,Csurka2008,shotton2008,Fidler2012}  confirm that doing global or local detection concurrently with segmentation can give significantly better segmentation results. In addition, \cite{Kontschieder2012,Rabinovich2007} incorporate the co-occurrence statistic as context information, to their detection and/or segmentation system, and showed that such context can improve accuracy. However, to the best of our knowledge, there is no work that trains using the two cues of a) the presence of the object in the image and b) such context information, together, to improve the segmentation process.

Our second contribution is to propose the Context Sensitive Image Level Prior (Context Sensitive ILP or ILPcont). Please note that, the terms context and co-occurrence will be used interchangeably from now on. From the experiment, Table~\ref{tab:result} shows that selecting the algorithm, which is aware of the co-occurrence of classes, to generate the Image Level Prior (ILP) produces much more promising results than the algorithm that trained to detect each class separately. The details of the Context Sensitive ILP are discussed next.

\subsubsection{Context Sensitive Image Level Prior}\label{ilp} Although Image Level Prior or image level class detection has been proved to be useful for semantic segmentation in recent papers, e.g. \cite{Boix2011,shotton2008}, previously the ILP only models the presence of classes for a single image. We propose a new image level class detection that takes into account co-occurrence statistics of the classes in the entire training data. The use of multi-label randomized trees allows us to model the global appearance of a single image and the co-occurrence statistics of the classes of entire training data at the same time. The multi-label randomized trees is first proposed in \cite{Dumont2009}, but it was used in different paradigm which is predicting a structured output. We, contrastingly, use the algorithm to learn both implicit class co-occurrence statistics of the training data and presences of classes in an image. Even though the algorithm is used to approach slightly different problems, the algorithm can be directly applied without any modification.

In this section, we will give a brief explanation of the multi-label randomized trees algorithm \cite{Dumont2009}. Multi-label random forests are random forests with a minor modification on the splitting quality metric. The metric is used to measure how well a feature splits the data at the node. The modification is made to take into account more than one class for the node splitting instead of a single class in the original randomized tree based model. The metric is based on the Gini entropy, and the modification are as follows, 

\begin{equation}\label{milp1}
\mathrm{score}(\mathbf{\mathcal{T}},\mathcal{F})=\frac{1}{C}\sum_{k=1}^{C}\mathrm{score}^k(\mathbf{\mathcal{T}},\mathcal{F}),
\end{equation}
\begin{equation}\label{milp2}
\mathrm{score}^k(\mathbf{\mathcal{T}},\mathcal{F})=G_k(\mathbf{\mathcal{T}})-G_{k|\mathcal{F}}(\mathbf{\mathcal{T}}),
\end{equation}
\begin{equation}\label{milp3}
G_k(\mathbf{\mathcal{T}})=2\big(\frac{\sum_{t_i\in \mathbf{\mathcal{T}}}\mathcal{L}_k(t_i)}{n}(1-\frac{\sum_{t_i\in \mathbf{\mathcal{T}}}\mathcal{L}_k(t_i)}{n})\big),
\end{equation}
\begin{equation}\label{milp4}
G_{k|\mathcal{F}}(\mathbf{\mathcal{T}})=G_k(\mathbf{\mathcal{T}_l})+G_k(\mathbf{\mathcal{T}_r}),
\end{equation}
where $\mathbf{\mathcal{T}}$ is data of size $n$ that reaches the node and $\mathcal{F}$ denotes a test function that routes subset of the data  $\mathbf{\mathcal{T}_l}$ to its left child node when all member of $\mathbf{\mathcal{T}_l}$ satisfy $\mathcal{F}$ otherwise routes the data $\mathbf{\mathcal{T}_r}$ to the right child node. $C$ is the number of semantic classes. $\mathcal{L}_k(t_i)$ is the function that returns 1 when $L_{ik}=1$, and 0 otherwise; and $L_{ik}$ indicates the presence of class $k$ in the datapoint $i^{th}$.


Table~\ref{tab:result} compares results of the original Semantic Texton Forests with our 2 proposed components. Please note that, combining the original STF with context sensitive ILP outperforms the full model of BoST that proposed in \cite{shotton2008}; our system has average recall 68.03\% compared to 66.9\% of \cite{shotton2008}.

\begin{table}[h]
\begin{center}
\begin{tabular}{|c|c|c|c|}
\hline
Method         & no ILP  & normal ILP & context ILP \\ \hline
STF (average)  & 31.02\% & 35.29\%$\ast$    & 68.03\%     \\ \hline
DSTF (average) & 39.67\% & 41.56\%    & \textbf{70.07\%$\dagger$}     \\ \hline
STF (global)   & 44.00\% & 50.16\%$\ast$    & 72.21\%     \\ \hline
DSTF (global)  & 49.82\% & 55.52\%    & \textbf{73.97\%$\dagger$}     \\ \hline
\end{tabular}
\end{center}
\caption{MSRC-21: Average and Global recalls of Decorrelated Semantic Texton Forests and Context Sensitive ILP ($\dagger$) and original Semantic Texton Forests and ILP (*).}
\label{tab:result}
\end{table}

\subsection{Location Potentials}

The last crucial ingredient are the location potentials. The location potentials are simply the statistics of how likely each absolute location in the image to be occupied by particular classes. The location potential is also used in \cite{shotton2007}. 

In this work, training images are first split into 2 groups:portrait images and landscape images. Next, for each group, we count the frequencies of each absolute location to be landed by a particular class. After having the location potentials for each class (each class has 2 location potentials: portrait and landscape), the location potentials are used as look up tables for an input location.

\begin{figure}[htb]
\begin{center}
   \includegraphics[width=1.0\linewidth]{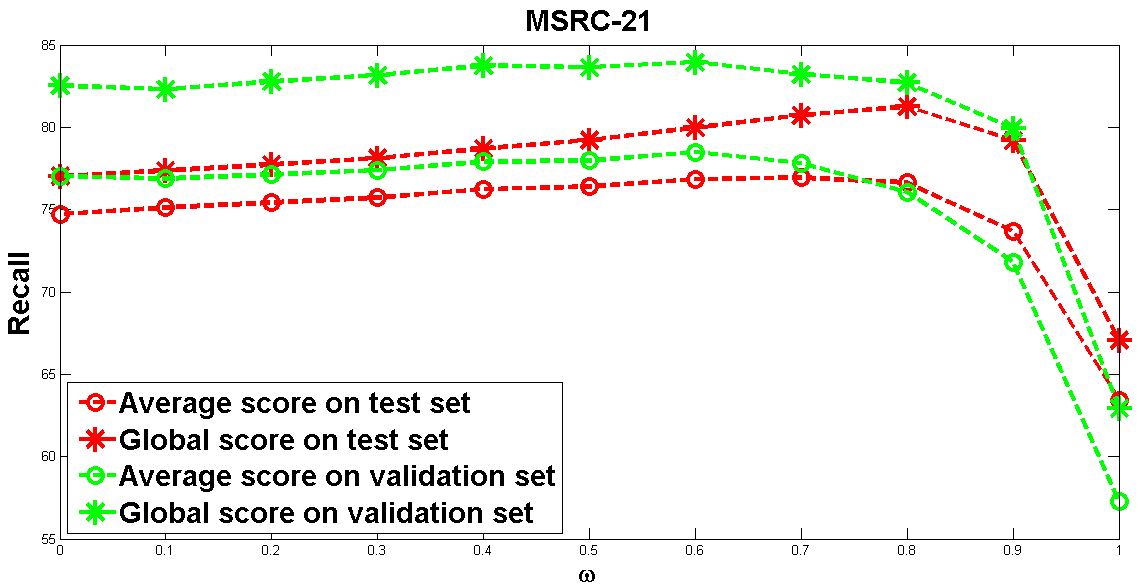}
\end{center}
   \caption{MSRC-21: Influence of location potentials, ranged from using pure DSTF, $\omega=0$, to pure location potentials, $\omega=1$ in the model. Note that, we optimise the system on average per-class recall.}
\label{fig:loc}
\end{figure}

Figure~\ref{fig:loc} illustrates the importance of the location potentials comparing to DSTF.

\subsection{Integration of the Components}

A simple Conditional Random Field model is selected to assemble the components. Simple, in our case, refers to utilising an ordinary grid graph, where each pixel has 4-connected neighbors, and the potts pairwise potential. This potential assigns low energy to two adjacent pixels with the same class, and high energy otherwise.

More formally, We cast the inference of our system as a minimization of the energy function 
\begin{equation}\label{eq:CRF}
 E(x_1,...,x_N|I)=\sum_{i\in N}\zeta_I\phi_i(x_i)+\sum_{(i,j)\in P}\psi_{i,j}(x_i,x_j),
\end{equation}
where $x_i$ is a random variable associated with the test image $I$. The random variable $x_i$ can be assigned one of the labels in the label set $\mathbf{c}=\{c_1,...,c_C\}$. $N$ is the number of pixels in the test image $I=\{x_1,...x_N\}$, and $P$ is a set of all pairs of neighboring pixels. $\psi$ is defined as the potts potential and $phi$is the unary potential defined as, 
\begin{equation}\label{eq:CRF2}
 \phi_i(x_i)=(1-\omega)*DSTF(x_i)+\omega*Location(x_i),
\end{equation}
and $\zeta_I$ is the image level prior of image $I$. Minimization is carried out using the graph cuts algorithm of Boykov~\etal\cite{Boykov2001a}.

 
Figure~\ref{fig:result} compares the average per-class recall results of our whole system, and when certain components are missing. The blue dotted line shows our results when all components were trained with standard data splitting, as per~\cite{shotton2007}. The red dotted line shows average per-class recalls of the system when only the ILP was trained with additional data, sampled from the test data, so the test data size is smaller than the standard one. The yellow dotted line shows the result when the ILP was trained in the unusual ways: the black star is the result when ILP was trained on the entire dataset, with no unseen data for the ILP, and the magenta star represents the result of Ideal ILP, assuming that we know the actual image tags.

\section{Results by Dataset}
We evaluate our approach via two well-known semantic segmentation datasets, MSRC-21 and PascalVOC-2010. MSRC-21 is now considered a small older dataset, but it is commonly used for validating semantic segmentation algorithms. Whereas the latter, PascalVOC-2010, is one of the newest and largest datasets. We choose these datasets to prove that our approach is robust with limited training data, as well as with great diversity of scenes. The main reason we select PascalVOC-2010 over the newer or the older versions of the same competition is the recently published finer ground associated with it~\cite{Mottaghi2014}.

\subsection{MSRC-21 \cite{shotton2007}}
The MSRC-21 dataset is composed of 591 images of size 320 x 213 and 213 x 320. The segmentation ground truth is made up of 21 classes which are mixed between background classes and object classes. The parameters we are using for the MSRC-21 dataset are the same as in \cite{shotton2008}, with one additional parameter $\omega$ to weight between the appearance potential (DSTF) and the location potential. We tune the extra parameter using the validation set as shown in Figure~\ref{fig:loc}.

\begin{figure}[htb]
\begin{center}

   \includegraphics[width=1.0\linewidth]{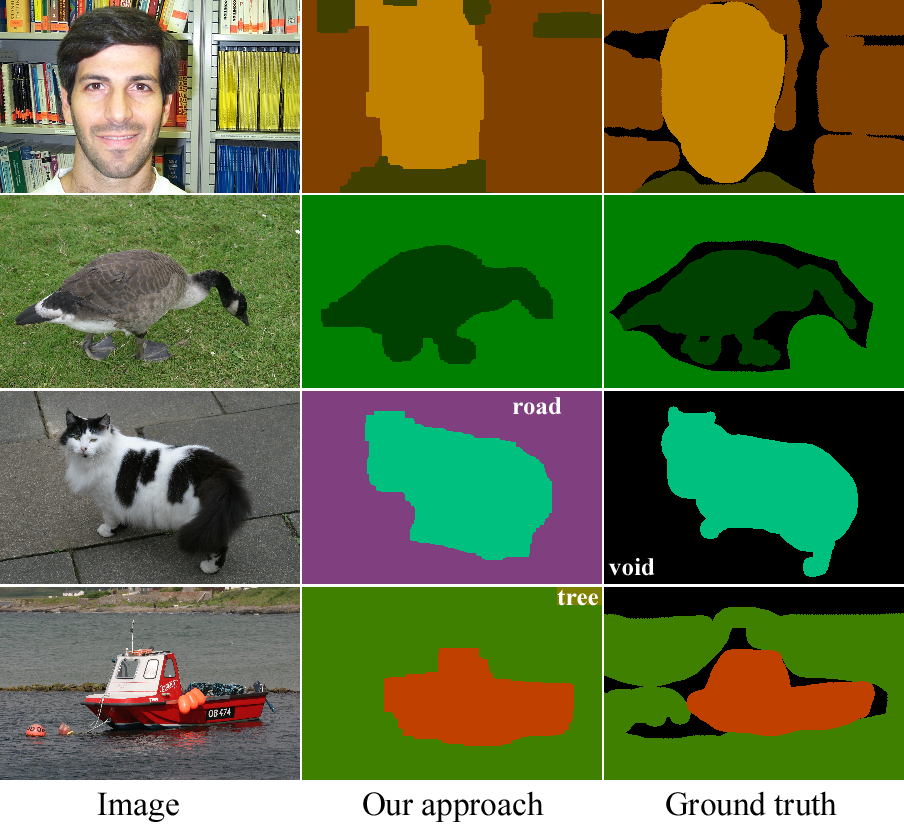}
\end{center}
   \caption{Qualitative results on the MSRC dataset.}
\label{fig:MSRCimg}
\end{figure}

\begin{figure*}[htb]
\begin{center}
   \includegraphics[width=0.9\linewidth]{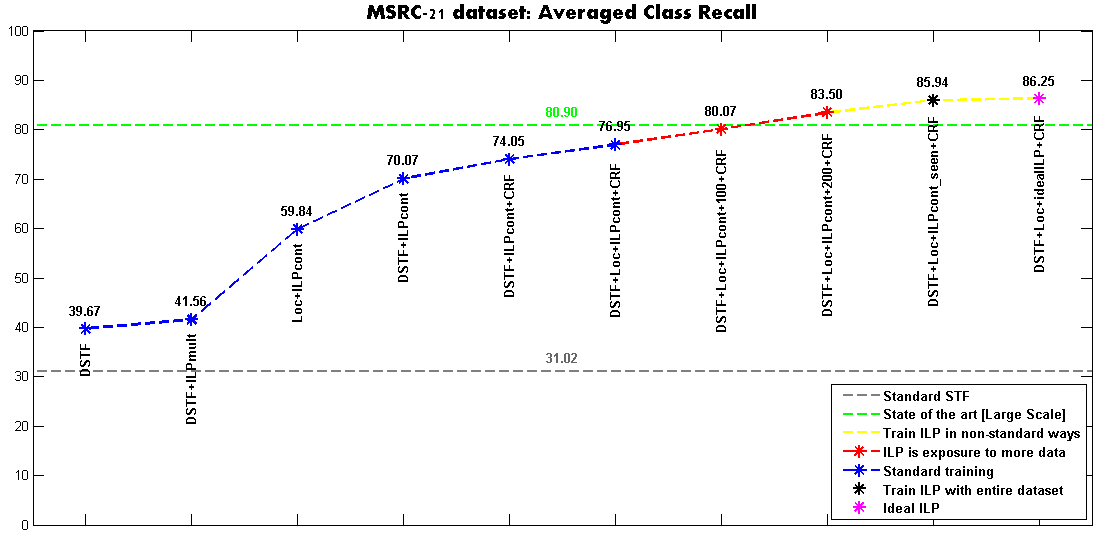}
\end{center}
   \caption{Average per-class recall results of the system (best viewed in colors) with or without certain components. The results are compared to the state of the art \cite{Alvarez2014}, green dotted line. Important notations: ILPmult (normal multi-class ILP regardless of the co-occurrence statistic), ILPcont (the proposed Context Sensitive ILP), ILPcont+N (the proposed Context Sensitive ILP when training with standard training and validation data + N images sampled from standard test data and testing on unseen data,therefore test data - N sampled images), ILPcont\_seen (the context sensitive ILP trained on all data; thus no unseen data for the ILP), and idealILP (When the ILP component produce 100\% correct prediction), and Loc (Location potentials).}
\label{fig:result}
\end{figure*}

Table~\ref{tab:result} demonstrates that both the proposed DSTF and context sensitive ILP work to  complement each other, and give $\approx$9\% and $\approx$37\% improvement respectively. Figure~\ref{fig:result} shows that our full model (integrating DSTF, Context Sensitive ILP, and Location potential by simple CRF) can achieve a result that is comparable to state of the art results. Besides, when the ILP has more training data, our proposed model even beats the top algorithm of this dataset. Table~\ref{tab:allresult} shows detailed results for each class compared to state of the art algorithms. One can observe that our system performs well on all classes. Qualitative results can be found in Fig.~\ref{fig:MSRCimg}.

\subsection{PascalVOC-2010 \cite{pascal-voc-2010}}
The PascalVOC-2010 semantic segmentation dataset consists of 964 training images, 964 validation images, and 964 test images. The dataset has 20 object classes and 1 background class which includes everything but the 20 object classes. The background class occupies 60.1\% of all pixels in the training and validation set~\cite{Mottaghi2014}. As the ground truth for the test set is not publicly available, the organizers run an evaluation server where users can submit up to two submissions per week. We test our algorithm on this dataset using default parameters, \ie the same parameters used for the MSRC-21 dataset. In addition, \cite{Mottaghi2014} relabeled the ground truth by adding more classes and removing the background class. We also show the results of training our ILP with this new ground truth data on the standard 20-object-classes, and with additional context classes, such as water, sky, road, etc. (+Add\_context). 

Table~\ref{tab:VOC2010} illustrates our quantitative results on the test data. Alone, STF and DSTF do not perform well with this dataset, since the data is more diverse and has a very large and complex background class. 
However, the Context Sensitive ILP still produces impressive improvements, improving the result by $\approx$17\% over the pure STF and DSTF, comparing to an improvement of only $\approx$7\% by using the multi-class ILP. Interestingly, our Context Sensitive ILP coupled with just Location potentials is proving very powerful, despite missing out on substantial information available to the full system. 

\begin{figure}[htb]
\begin{center}

   \includegraphics[width=1.0\linewidth]{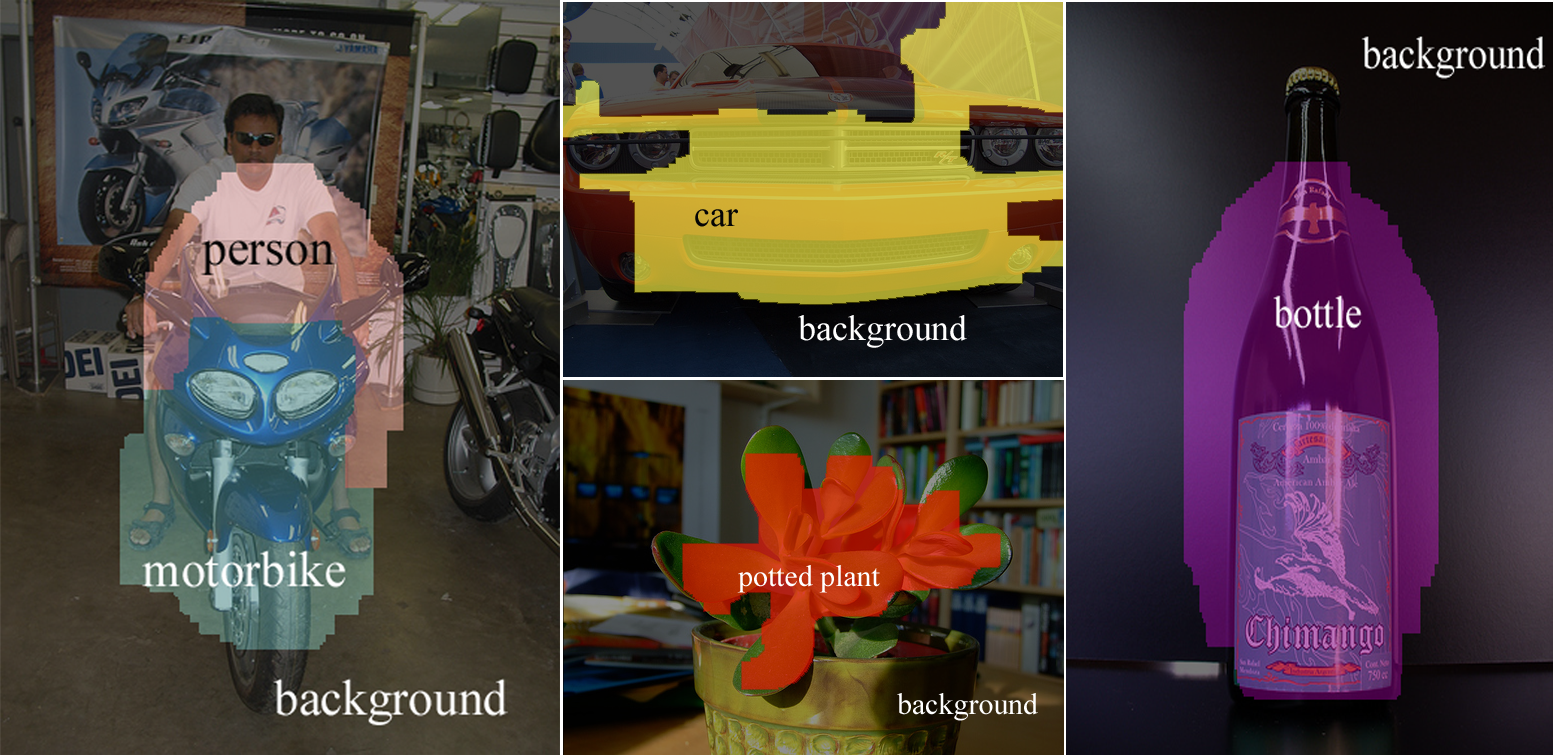}
\end{center}
   \caption{Qualitative results on the PascalVOC-2010 dataset when using our approach. We illustrates the results overlayed on the image since the ground truth for is not available.}
\label{fig:VOCimg}
\end{figure}

Although the new ground truth on VOC has better ground truth, we can see that the accuracy decreases. The relabeling  process has modified the ground truth for the standard 20-object-classes, but we still evaluate the result via the evaluation server which evaluates based on the old ground truth. Furthermore, adding more context classes can hurt the accuracy of the system because a larger number of classes reduces the ILP prediction accuracy. Fig.~\ref{fig:VOCimg} demonstrates some qualitative results.

\begin{table}[h]
\begin{center}
\begin{tabular}{|l|c|}\hline
Methods                            & IoU \\ \hline
\underline{Train on original ground truth~\cite{pascal-voc-2010}}&\\
STF                        & 1.454   \\
DSTF						& 0.858    \\
STF+ILPmult                & 6.019$\ddagger$   \\
STF+ILPcont                  & 16.656$\ddagger$ \\
DSTF+ILPcont                 & 16.947$\ddagger$ \\
Location+ILPcont              & 21.474$\star$ \\
STF+Loc+ILPmult                  & 7.060$\ddagger$  \\
STF+Loc+ILPcont                & 21.403 $\ddagger$ \\
\textbf{Our full model} (DSTF+Loc+ILPcont)     &  21.588$\ddagger$ \\
\textbf{Our full model} (DSTF+Loc+ILPcont)     &  24.058$\star$  \\\hline 
\underline{Train ILP on the new ground truth \cite{Mottaghi2014}}&\\
STF+Loc+ILPmult                & 6.585$\ddagger$   \\
STF+Loc+ILPcont                 & 17.760$\ddagger$  \\
STF+Loc+ILPcont+Add\_context                  & 16.875$\ddagger$  \\
\underline{State of the art algorithms}&\\
Topic model \cite{Gonzalez-Diaz2013}  	&  27.8 \\
DenseCRF \cite{Krahenbuhl2011} (non-standard test set) & 30.2 \\
HCRF+Cooc \cite{Ladicky2010}   & 30.3 \\
Whole \cite{Fidler2012} (non-standard test set)      &  31.2 \\
Harmony$_4$ \cite{Boix2011}      &  38.0 \\
Composite \cite{Li2013}     &  49.6 \\\hline   
\end{tabular}
\end{center}
\caption{Intersection over Union score of the system on PascalVOC-2010 dataset. We demonstrate the results from different combinations of our proposed components. Please note that, ILPmult and ILPcont represent the multi-class image level prior and context sensitive image level prior respectively. Loc stands for location potential, and Add\_context represents us training the ILP with extra context classes: sky, road, building, water, grass. $\ddagger$ indicates that the methods use that same set of parameters, to make the numbers comparable; $\star$ indicates the parameter was tuned by cross validating, fixing the ILP for class background to probability 0.1, 0.2, ..., 1.0.}
\label{tab:VOC2010}
\end{table}

\section{Limitations}

We validated our proposed system on another standard dataset, that demonstrates a predictable limitation of our approach. The CamVid dataset~\cite{Brostow2008} consists of image sequences of road scenes, where the ground truth labels associate each pixel with one of the grouped 11 semantic classes. To be comparable to other algorithms, we downsample all the images by a factor of 3 as in \cite{Sturgess2010}. 

Table~\ref{tab:CamVid} shows the quantitative results of our algorithm on the CamVid dataset. Since most of the images have almost the same set of classes present in them, context information is not useful here. Therefore, we can see that the Context Sensitive ILP performed worse than the normal multi-class ILP because the context sensitive ILP can extract very few co-occurence patterns from the training data. DSTF also hurts accuracy because the training sub-STFs are not really specialized, \ie they have access to artificially small training sets, with little difference between scene categories.

\begin{table}[h]
\begin{center}
\begin{tabular}{|l|c|c|}\hline
Methods                           & Average & Global \\ \hline
STF                        & 29.95   & 27.25  \\
STF+ILPcont               & 27.31   & 27.20  \\
STF+ILPmult              & 29.53   & 29.38  \\
DSTF                        & 10.41   & 7.38  \\
DSTF+ILPmult             & 26.84   & 28.32  \\
Loc+ILPmult              & 27.85   & 55.96  \\
STF+Loc+ILPmult  & \textbf{40.27}   & \textbf{59.39}  \\
\textbf{Full model} (DSTF+Loc+ILPmult) & 34.56     &  52.23 \\ \hline
\underline{State of the art algorithm} & & \\ 
Combining Object Detection~\cite{Sturgess2010} & 62.5 & 83.8 \\ \hline
\end{tabular}

\end{center}
\caption{Average and Global recalls of the system on CamVid dataset. We tested our proposed model on different combinations of the components.}
\label{tab:CamVid}
\end{table}

\section{Conclusion}
We have shown that a combination of simple techniques can yield excellent accuracy, given only a limited computational budget. Our DSTF shows an impressive ability to empower the inaccurate appearance predictions of a normal STF, with only a small extra overhead. This is noteworthy becaus each sub-STF is working with less training data. The Context Sensitive ILP proved quite capable of recovering from even fairly bad appearance predictions. While other ILP models have been proposed previously, using the co-occurrence statistics jointly with image level class detection can now be accomplished cheaply, and can yield a substantial improvement in accuracy.

We are making our code publicly available. Many extensions for the future are possible because the existing system is simple and complementary to many other approaches. A natural extension would use fast filters over the image as extra appearance channels in the DSTF. It could also be fruitful to learn a variety of location potentials, \ie for different camera poses, \eg from car-mounted or hand-held cameras.


\begin{table*}[tb]
\scriptsize
\begin{center}
\begin{tabular}{llllllllllllllllllllllll}
\hline
                  & \rotatebox{90}{building} & \rotatebox{90}{tree}  & \rotatebox{90}{road}  & \rotatebox{90}{cow}   & \rotatebox{90}{grass} & \rotatebox{90}{sky}   & \rotatebox{90}{sheep} & \rotatebox{90}{aeroplane}& \rotatebox{90}{body}  & \rotatebox{90}{face}  & \rotatebox{90}{book}  & \rotatebox{90}{water} & \rotatebox{90}{car}   & \rotatebox{90}{bicycle} & \rotatebox{90}{flower} & \rotatebox{90}{sign}  & \rotatebox{90}{bird}  & \rotatebox{90}{chair} & \rotatebox{90}{cat}   & \rotatebox{90}{dog}   & \rotatebox{90}{boat}  & \rotatebox{90}{Average} & \rotatebox{90}{Global} \\ \hline
BoST \cite{shotton2008}           & 49       & 79    & 78    & \textbf{97}    & 88    & 78    & \textbf{97}    & 82        & 66    & 87    & 93    & 54    & 74    & 72      & 74     & 36    & 24    & 51    & 75    & 35    & 18    & 67      & 72     \\ \hline
Harmony$_1$ \cite{Gonfaus2010}		& 60     & 77  & 76  & 91   & 78  & 88  & 68   & 87      & 56  & 73  & 95  & 76& 77  & 93    & 97   & 73  & 57  & 81  & 81  & 46  & 46  & 75    & 77    \\ \hline
HCRF+Cooc \cite{Ladicky2010}         & \textbf{82}       & 88    & \textbf{93}    & 73    & 95    & \textbf{100}    & 88    & 83        & 65    & 88    & 85    & \textbf{92}    & 87    & 88      & 96     & 96    & 27    & 37    & 49    & 80    & 20    & 77    & 87   \\ \hline
DenseCRF \cite{Krahenbuhl2011}          & 75       & \textbf{91}    & 90    & 84    & \textbf{99}    & 95    & 82    & 82        & 80    & 89    & 98    & 71    & \textbf{90}    & 94      & 95     & 77    & 48    & 61    & 78    & 48    & 22    & 78   & 86     \\ \hline
Whole \cite{Fidler2012} & 71       & 90    & 89    & 79    & 98    & 93    & 86    & \textbf{88}        & 68    & 90    & 97    & 86    & 84    & 94      & 98     & 76    & 53    & 71    & 83    & 55    & 17    & 79    & 86   \\ \hline
Harmony$_4$ \cite{Boix2011}        & 66       & 84    & 82    & 81    & 87    & 93    & 83    & 81   & 70    & 78    & 90    & 82    & 86    & 94      & 96     & 87    & 48    & 81    & 82    & 75    & 52    & 80    & 83   \\ \hline
Large Scale \cite{Alvarez2014}     & 73       & 90    & 90    & 85    & \textbf{99}    & 95    & 82    & 86        & \textbf{87}    & 91    & 96    & 74    & 88    & 91      & 96     & 83    & 54    & 79    & 81    & 60    & 18    & 81    & 87   \\ \hline
\rowcolor[HTML]{FFCE93} 
Our               & 51    & 79 & 85 & 92 & 96 & 81 & 90  & 68     & 59 & \textbf{93} & 95 & 84 & 76 & 92   & 98  & 75 & 64 & 71 & \textbf{86} & 47 & 34 & 77   & 80  \\ \hline
\rowcolor[HTML]{FFCE93} 
Our+100     & 58    & 86 & 89 & 92& 97 & 87 & 93 & 67    & 61 & 90 & 94 & 87 & 75 & 92    & \textbf{100}    & 88 & 70 & 83 & 77 & \textbf{91} & \textbf{70} & 83   & 85  \\ \hline 
\rowcolor[HTML]{FFCE93} 
Our+200     & 71   & 86 & 88    & 94 & 98 & 86 & 95 & 79     & 68    & 83 & \textbf{99} & 86 & 76 & \textbf{96}   & \textbf{100}    & \textbf{100} & \textbf{91} & \textbf{94} & 80 & 85 & 51 & \textbf{86}  & \textbf{88} \\ \hline \hline
\rowcolor[HTML]{F5A9BC} 
Our+Seen       & 68   & 88 & 87 & 94 & 97 & 89 & 94 & 71 & 60 & 92 & 99 & 92 & 85 & 93   & 98  & 93 & 78 & 96 & 89 & 81 & 61 & 86   & 88  \\ \hline
\rowcolor[HTML]{F5A9BC} 
Our+Ideal    & 71   & 88 & 87 & 94 & 96 & 90 & 95 & 70 & 59 & 91  & 99 & 91 & 83 & 89   & 98  & 92  & 79& 97  & 90 & 81 & 71 & 86  &88  \\ \hline
Harmony$_4$+Ideal \cite{Boix2011}   & 68   & 92 & 89 & 86 & 93 & 97 & 88 & 91 & 60 & 73 & 100 & 85 & 86 & 94   & 100  & 89  & 77 & 96  & 95 & 94 & 74 & 87  &89  \\ \hline
\rowcolor[HTML]{F5A9BC}

\end{tabular}
\end{center}
\caption{MSRC-21 segmentation results. Note that we show results of our system and \cite{Boix2011} with Seen ILP and Ideal ILP to show the upper bound of the systems thus we do not include them to the comparison.}
\label{tab:allresult}
\end{table*}

{\small
\bibliographystyle{ieee}
\bibliography{CVPR15}

\begin{thebibliography}{10}\itemsep=-1pt

\bibitem{Alvarez2014}
J.~M. Alvarez, M.~Salzmann, and N.~Barnes.
\newblock {Large-scale semantic co-labeling of image sets}.
\newblock In {\em IEEE Winter Conference on Applications of Computer Vision},
  pages 501--508, Mar. 2014.

\bibitem{Boix2011}
X.~Boix, J.~M. Gonfaus, J.~van~de Weijer, A.~D. Bagdanov, J.~Serrat, and
  J.~Gonz\`{a}lez.
\newblock {Harmony Potentials Fusing Global and Local Scale for Semantic Image
  Segmentation}.
\newblock {\em International Journal of Computer Vision}, 96(1):83--102, Apr.
  2011.

\bibitem{Boykov2001a}
Y.~Boykov, O.~Veksler, and R.~Zabih.
\newblock {Fast Approximate Energy Minimization via Graph Cuts}.
\newblock {\em Pattern Analysis and Machine Intelligence 2001},
  (November):1222--1239, 2001.

\bibitem{Brostow2008}
G.~J. Brostow, J.~Shotton, J.~Fauqueur, and R.~Cipolla.
\newblock {Segmentation and Recognition using Structure from Motion Point
  Clouds}.
\newblock In {\em ECCV}, pages 1--14, 2008.

\bibitem{Chatfield2014}
K.~Chatfield, K.~Simonyan, A.~Vedaldi, and A.~Zisserman.
\newblock {Return of the Devil in the Details : Delving Deep into Convolutional
  Nets}.
\newblock In {\em BMVC}, 2014.

\bibitem{Couprie2013a}
C.~Couprie, C.~Farabet, L.~Najman, and Y.~LeCun.
\newblock {Indoor Semantic Segmentation using depth information}.
\newblock In {\em International Conference on Learning Representation}, pages
  1--7, 2013.

\bibitem{Csurka2008}
G.~Csurka and F.~Perronnin.
\newblock {A Simple High Performance Approach to Semantic Segmentation}.
\newblock In {\em BMVC}, pages 22.1--22.10. British Machine Vision Association,
  2008.

\bibitem{Dumont2009}
M.~Dumont, R.~Maree, L.~Wehenkel, and P.~Geurts.
\newblock {Fast Multi-class Image Annotation with Random Subwindows and
  Multiple Output Randomized Trees}.
\newblock In {\em VISAPP}, 2009.

\bibitem{pascal-voc-2010}
M.~Everingham, L.~Van~Gool, C.~K.~I. Williams, J.~Winn, and A.~Zisserman.
\newblock The {PASCAL} {V}isual {O}bject {C}lasses {C}hallenge 2010 {(VOC2010)}
  {R}esults.
\newblock
  http://www.pascal-network.org/challenges/VOC/voc2010/workshop/index.html.

\bibitem{farabet2013}
C.~Farabet, C.~Couprie, L.~Najman, and Y.~LeCun.
\newblock {Learning Hierarchical Features for Scene Labeling}.
\newblock {\em PAMI}, pages 1--15, 2013.

\bibitem{Gonfaus2010}
J.~M. Gonfaus, X.~Boix, J.~van~de Weijer, A.~D. Bagdanov, J.~Serrat, and
  J.~Gonzalez.
\newblock {Harmony Potentials for Joint Classification and Segmentation}.
\newblock {\em CVPR}, pages 3280--3287, June 2010.

\bibitem{Gonzalez-Diaz2013}
I.~Gonz\'{a}lez-D\'{\i}az and F.~D\'{\i}az-de Mar\'{\i}a.
\newblock {A region-centered topic model for object discovery and
  category-based image segmentation}.
\newblock {\em Pattern Recognition}, 46(9):2437--2449, Sept. 2013.

\bibitem{Hariharan2014}
B.~Hariharan, P.~Arbel\'{a}ez, R.~Girshick, and J.~Malik.
\newblock {Simultaneous Detection and Segmentation}.
\newblock In {\em ECCV}, pages 1--16, 2014.

\bibitem{Kontschieder2012}
P.~Kontschieder, S.~Rota~Bulo, A.~Criminisi, P.~Kohli, M.~Pelillo, and
  H.~Bischof.
\newblock {Context-Sensitive Decision Forests for Object Detection}.
\newblock In {\em NIPS}, pages 1--9, 2012.

\bibitem{Krahenbuhl2011}
P.~Kr\"{a}henb\"{u}hl and V.~Koltun.
\newblock {Efficient inference in fully connected crfs with gaussian edge
  potentials}.
\newblock In {\em NIPS}, pages 1--9, 2011.

\bibitem{Ladick2009}
L.~Ladicky, C.~Russell, P.~Kohli, and P.~H.~S. Torr.
\newblock {Associative Hierarchical CRFs for Object Class Image Segmentation}.
\newblock In {\em ICCV}, 2009.

\bibitem{Ladicky2010}
L.~Ladicky, C.~Russell, P.~Kohli, and P.~H.~S. Torr.
\newblock {Graph Cut Based Inference with Co-occurrence Statistics}.
\newblock In {\em ECCV}, pages 1--14, 2010.

\bibitem{Sturgess2010}
L.~Ladicky, P.~Sturgess, K.~Alahari, C.~Russell, and P.~H.~S. Torr.
\newblock {What , Where \& How Many ? Combining Object Detectors and CRFs}.
\newblock In {\em ECCV}, 2010.

\bibitem{Li2013}
F.~Li, J.~Carreira, G.~Lebanon, and C.~Sminchisescu.
\newblock {Composite Statistical Inference for Semantic Segmentation}.
\newblock In {\em CVPR}, number~1, pages 3302--3309. Ieee, June 2013.

\bibitem{Mottaghi2014}
R.~Mottaghi, X.~Chen, X.~Liu, N.~Cho, S.~Lee, S.~Fidler, R.~Urtasun, and
  A.~Yuille.
\newblock {The Role of Context for Object Detection and Semantic Segmentation
  in the Wild}.
\newblock In {\em CVPR}, 2014.

\bibitem{Rabinovich2007}
A.~Rabinovich, A.~Vedaldi, C.~Galleguillos, E.~Wiewiora, and S.~Belongie.
\newblock {Objects in Context}.
\newblock In {\em ICCV}, pages 1--8. Ieee, 2007.

\bibitem{shotton2008}
J.~Shotton, M.~Johnson, and R.~Cipolla.
\newblock {Semantic texton forests for image categorization and segmentation}.
\newblock In {\em CVPR}, pages 1--8, June 2008.

\bibitem{shotton2007}
J.~Shotton, J.~Winn, C.~Rother, and A.~Criminisi.
\newblock {TextonBoost for Image Understanding: Multi-Class Object Recognition
  and Segmentation by Jointly Modeling Texture, Layout, and Context}.
\newblock {\em International Journal of Computer Vision}, 81(1):2--23, Dec.
  2007.

\bibitem{Fidler2012}
J.~Yao, S.~Fidler, and R.~Urtasun.
\newblock {Describing the scene as a whole: Joint object detection, scene
  classification and semantic segmentation}.
\newblock In {\em CVPR}, pages 702--709, June 2012.

\end{thebibliography}
}

\end{document}